\newif\ifarxiv
\arxivtrue 

\ifarxiv
	\documentclass[12pt]{article}
	\usepackage[margin=1in]{geometry}
	\usepackage[slantedGreek]{mathpazo}
	\usepackage[pdftex]{graphicx}
	\usepackage{amsfonts}
	\usepackage{amssymb}
	\usepackage{amsthm}
	\usepackage{graphicx,float}
	\usepackage{amsmath}%
    \usepackage{subcaption} 
    \usepackage{natbib} 
	\usepackage{hyperref}
    \usepackage{accents}
\else
    \usepackage{natbib} 
    \usepackage{hyperref}
\fi

\usepackage{color,soul,marginnote}

\newcommand{\E}[1]{\mathbb{E}\left[#1\right]} 
\newcommand{\Ewrt}[2]{\mathbb{E}_{#1}\left[#2\right]} 

\newcommand{\bx}{{\bf x}}

\DeclareMathOperator*{\argmin}{arg\,min} 

\newtheorem{theorem}{Theorem}[section]

\newcommand{\vnorm}[1]{\left|\left|#1\right|\right|}
\ifarxiv
\title{Bayesian Double Descent}
\author{	
    \makebox[.4\linewidth]{Nick Polson}\\\textit{Booth School of Business}\\\textit{University of Chicago}\\\and 
    \makebox[.4\linewidth]{Vadim Sokolov\footnote{Nick Polson is at Chicago Booth: ngp@chicagobooth.edu. Vadim Sokolov is Associate Professor at Volgenau School of Engineering, George Mason University, USA: vsokolov@gmu.edu.}}\\\textit{Department of Systems Engineering}\\\textit{and Operations Research}\\\textit{George Mason University}
}
\date{First Draft: December 25, 2024\\This Draft: \today}
\else
\fi

\graphicspath{{./fig/}}

\begin{document}
\ifarxiv
\maketitle

\begin{quote}
    \emph{"It is vain to do with more what could be done with less"------}
    \textit{William of Ockham (c. 1285--1347)}
\end{quote}

\begin{abstract}
\noindent	Double descent is a phenomenon of over-parameterized statistical models such as deep neural networks which have a re-descending property in their risk function.  As the complexity of the model increases, risk
exhibits a U-shaped region due to the traditional bias-variance trade-off, then as the number of parameters equals the number of observations and the model becomes one of interpolation where the risk can be unbounded and finally, in the over-parameterized region, it re-descends---the double descent effect. Our goal is to show that this has a natural Bayesian interpretation. We also show that this is not in conflict with the traditional Occam's razor---simpler models are preferred to complex ones, all else being equal. 
Our theoretical foundations use Bayesian model selection, the Dickey-Savage density ratio, and connect generalized ridge regression and global-local shrinkage methods with double descent. We illustrate our approach for high dimensional neural networks and provide detailed treatments of infinite Gaussian means models and non-parametric regression. Finally, we conclude with directions for future research.
\end{abstract}
\else
\fi


\newpage
\section{Introduction}\label{sec:intro}

Double descent---a phenomenon of over-parameterized models---is the property that an estimator's risk re-descends as the number of parameters extends past the interpolation limit. Empirically, the double descent effect was initially observed for high-dimensional neural network regression models and the good performance of deep learning on such tasks as large language models, image processing, and generative AI methods. Bias-variance trade-off occurs on the risk as a property of shrinkage estimators. Double descent extends the classical bias-variance tradeoff associated with modern day machine learning methods. 
This phenomenon was first observed in the context of linear regression~\cite{belkin2019reconciling}, where the authors showed that the test error of the estimator can decrease as the number of parameters increases. \cite{bach2024highdimensional} extends these results to stochastic regression models. 

Occam's razor---the favoring of simpler models over complex ones---is a natural feature of Bayesian methods which base their inference for model selection on the weight of evidence (a.k.a. the marginal likelihood of the data). In doing so, Bayes penalizes models with higher complexity via a correction term as in the Bayesian Information Criterion (BIC). This seems incompatible with the double descent phenomenon. We show this is not the case, as even though Bayesian methods shift the posterior towards lower-complexity models, highly parameterized Bayesian models can also have good risk properties due to the conditional prior of high dimensional parameters given the model. We illustrate this with an application to neural network models.

Interpolators---estimators that achieve zero training error---were then shown to have attractive properties due to the double descent effect~\cite{hastie2022surprises}. Our goal is to show that Bayesian estimators can also possess a double descent phenomenon. Interpolators such as ReLU neural networks~\cite{polson2017deep} have increased in popularity with many applications such as traffic flow modeling~\cite{polson2017deep} and high-frequency trading~\cite{dixon2019deep}, among many others.

\begin{figure}[H]
	\centering
	\includegraphics[width=0.5\textwidth]{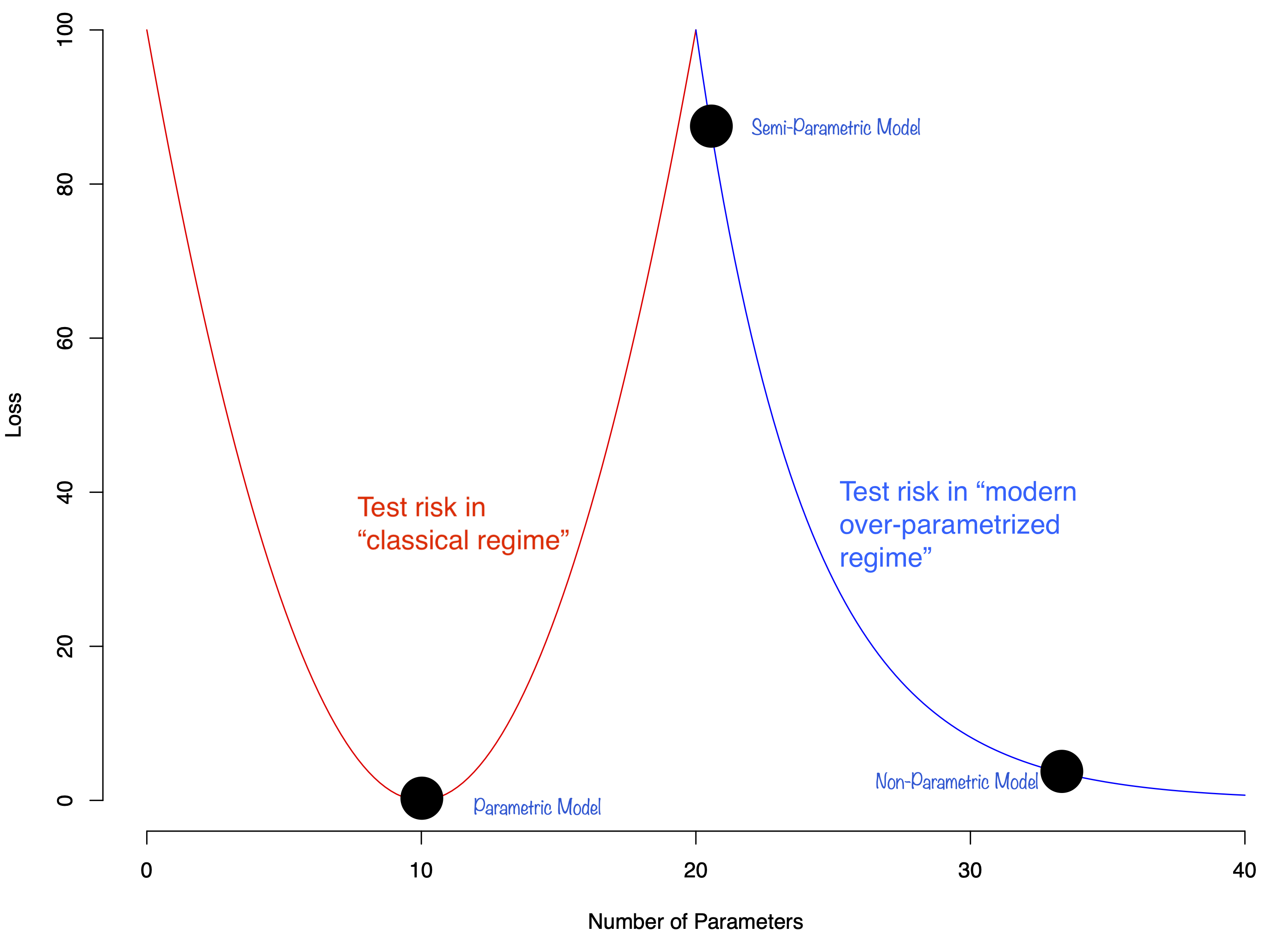}
	\caption{Stylized double descent curve showing the classical bias-variance trade-off region (left) and the re-descending risk in the over-parameterized regime (right). The interpolation threshold occurs when the number of parameters equals the number of observations.}\label{fig:double-descent}
\end{figure}

Double descent has been studied from a frequentist point of view in \cite{belkin2019reconciling,bach2024highdimensional}. 
The phenomenon of double descent is illustrated in Figure~\ref{fig:double-descent}. The first part of the curve represents the classical U-shaped bias-variance trade-off. The second part demonstrates the double descent phenomenon, where the test error of the estimator can decrease as the model becomes over-parameterized beyond the interpolation threshold. This phenomenon was later observed in the context of deep learning~\cite{nakkiran2021deep}. The authors showed that the test error of the estimator can decrease as the number of parameters increases.
From a Bayesian perspective, we show that there is always Occam's razor---the marginal likelihood (being an integral) will always favor lower-dimensional models over higher-dimensional ones.
There is always a bias-variance trade-off in the construction of Bayesian estimators. We now formalize this with a discussion of BIC.

The rest of this paper is outlined as follows. The next subsection provides connections with the existing literature on bias-variance trade-offs. Section~\ref{sec:bayesian-model-complexity} provides a framework for inference on model complexity and parameters, together with defining Bayes risk, which is an essential part of understanding double descent. Section~\ref{sec:bayesian-double-descent} provides a Bayesian definition of double descent. Section~\ref{sec:bayesian-interpolation} studies the double descent phenomenon in a model for Bayesian interpolation. Section~\ref{sec:nn-regression} provides an application to Bayesian interpolation and neural network regression. Section~\ref{sec:grr-shrinkage} provides comprehensive coverage of generalized ridge regression and shrinkage methods, including infinite Gaussian means and non-parametric regression. See also~\cite{mackay1992bayesian}, who provides a discussion of the weight of evidence and its relationship to model selection and hyperparameter regularization. Finally, Section~\ref{sec:discussion} concludes with directions for future research.

\subsection{Bias-Variance Trade-off in Neural Networks}
\cite{geman1992neural} were the first to explicitly mention a bias-variance trade-off. From a statistical viewpoint, they observed that a feed-forward NN (trained by backprop) is an example of non-parametric regression. As \cite{geman1992neural} pointed out, the variance is generally reduced by smoothing at the expense of missing peaks and valleys in the true function.  \cite{white1992nonparametric} used sieve type methods and provide a conditional quantile NN regression framework. See also \cite{barron1993universal}. Although, there exist many consistent non-parametric estimators, but the choice of estimator class remains an open question. Should we use quantile neural networks (QNNs), ReLU networks, kernel methods, or other approaches? This remains an active area of research, with the optimal choice often being problem-specific and dependent on the underlying data structure and computational constraints.
  
Let $D = \{(y_i, \bx_i)\}_{i=1}^N$ be an observed set of input-output pairs. We consider the case of stochastic regressors. The goal is to predict a new $y = f(\bx) + \epsilon $  for a new $ \bx $.  Under squared error loss the optimal predictor is the conditional mean $\E{y| \bx}$.  The risk is then given by
$$
\Ewrt{ y , \bx, D}{( y - f( \bx , D ) )^2 \mid \bx , D } 
$$
And this has the usual bias-variance decomposition. 
$$
\Ewrt{D}{( f ( \bx , D ) - \E{y\mid \bx} )^2} = {\rm bias} + {\rm variance }
$$
From a Bayesian perspective, methods are admissible (and never unbiased) and they dominate classical procedures by introducing a small bias in exchange for a significant reduction in variance. Double descent from a Bayesian perspective is just the insight that the variance term can still be large. The bias always comes from the prior, and more careful prior specifications can reduce the variance (although no proper Bayes rule can be dominate in risk).  

\cite{mackay1992bayesian} warns that training and test error rates are likely to be "uncorrelated". The double descent property may be partly a mirage and a consequence of \emph{a priori} regularization specifications. Interpolation of training data is simply a property of a highly non-parametric model. As we demonstrate in this paper, empirical risk bounds that approximate true Bayes risk are of limited use for understanding double descent, as are functional class complexity measures such as VC-dimension, because they disregard the crucial role of prior regularization.

One cannot get away from the fact that the Bayes rule is the optimal processing rule~\cite{akaike1974new}, and the predictive mean is simply the optimal rule. \cite{rubin1988robustness} provides an interesting perspective on Bayes robustness. \cite{stone1974crossvalidatory} points to the inconsistencies in selecting hyper-parameters based on cross-validation.

To illustrate the double descent phenomenon in a concrete setting, we present a detailed example using polynomial regression with Legendre basis functions. This example demonstrates how the test error can exhibit the characteristic U-shaped curve followed by a re-descent as model complexity increases far beyond the interpolation threshold.
Young argues that the best approach to prediction using polynomials is to fit the largest possible degree commensurate with our computing and statistical skills. Savage argues that a model should be as big as an elephant---although four parameters is ample to define an elephant! We formalise these thoughts in a Bayesian model selection framework. Our main result is that, under suitable compatibility conditions for the priors given models, we can calculate the posterior model probabilities and parameter posteriors under the simpler model from posterior ordinates under the complex model. Hence, in double descent, we should use the posterior estimator that we have calculated from the over-parameterised model only. A caveat is that we have to be able to specify joint parameter priors over complex spaces---an ongoing challenge for the Bayesian paradigm. We illustrate the Bayesian double descent phenomenon for polynomial regression with global-local priors \cite{polson2012local}.

\section{Bayesian Model Complexity}\label{sec:bayesian-model-complexity}

\begin{quote}
\emph{Models should be simple but no simpler} (Attributed to Albert Einstein)
\end{quote}

Throughout this paper, we use $\boldsymbol{\theta}_M$ to denote parameter vectors, $M$ and $m$ to denote "complex" and "simple" models correspondingly, $D$ to denote data, and $p(\cdot)$ for probability densities. The conditional prior $p(\boldsymbol{\theta}_M | M)$ plays a crucial role in our analysis of double descent. We use $\mathcal{M}$ to denote the space of models, $\Theta_M$ to denote the parameter space for model $M$, and $n$ to denote the sample size. For functions, we use $f(\cdot)$ to denote the true function and $\hat f(\cdot)$ to denote estimators.

The Bayesian paradigm provides a coherent framework to simultaneously infer parameters and model complexity. Double descent (a.k.a. the re-descending nature of Bayes risk as a function of complexity) can then be studied through a Bayesian lens. The optimal predictive rule is adaptive in the sense that it is a weighted average of individual predictors given complexity. 

Our approach shows that double descent is intimately related to the \emph{a priori} specification of the parameter distribution given complexity. While this acts as an automatic Occam's razor via the marginal likelihood of model complexity, classical measures such as VC-dimension and asymptotic rates of convergence disregard prior (a.k.a. regularization penalties) and as such miss much of the action in the double descent risk.
Let $D$ denote data and $\boldsymbol{\theta}_M \in \Theta_M$ denote a set of parameters under model $M \in \mathcal{M}$. 
Given $\boldsymbol{\theta}_M = (\theta_1, \ldots, \theta_M)$ a vector of parameters, the Bayesian approach is straightforward: implement the Bayesian paradigm by executing Bayes rule. This requires the laws of probability and not optimization techniques. The notion of model complexity is no different. Let $\mathcal{M}$ denote the space of models. The Bayesian paradigm simply places probabilities over parameters and models given the data, namely $p(\boldsymbol{\theta}_M, M \mid D)$.

This has a number of decompositions. Bayes theorem calculates the joint posterior over parameters and models given data $D$, namely
\[
    P(\boldsymbol{\theta}_M,M\mid D) = P(\boldsymbol{\theta}_M \mid M,D)P(M\mid D).
\]
Notice how this factors the posterior into two terms: the conditional posterior over parameters given the model and the posterior over models given data.

\paragraph{Evidence given $M$} The key quantity is the weight of evidence (a.k.a. marginal distribution of the data $D$ given the model $M$), defined by
\[
p( D | M ) = \int_{ \Theta_M } p( D \mid \boldsymbol{\theta}_M , M )  p ( \boldsymbol{\theta}_M | M ) d \boldsymbol{\theta}_M.
\]
Here $p( D \mid \boldsymbol{\theta}_M , M )$ is the traditional likelihood function. 

The key conditional distribution, however, is the specification of the prior over parameters $p(\boldsymbol{\theta}_M | M)$.
As this is used in the marginalization, it can affect the Bayes risk dramatically. Occam's razor comes from the fact that this marginalization provides a weight of evidence that favors simpler models over
more complex ones.  We will argue that  double descent, from a Bayesian perspective, is governed by the prior of
parameters given the model, namely $p(\boldsymbol{\theta}_M | M)$.

More precisely, we will see that the prior $p(\boldsymbol{\theta}_M | M)$ will lead to an Occam's razor effect, namely that the marginal distribution will favor simpler models. This will be quantified by the Bayesian Information Criterion (BIC). Importantly, this Occam's razor effect is not in conflict with the Bayesian double descent phenomenon, which emerges from the marginal posterior of models given data and the conditional prior specification $p(\boldsymbol{\theta}_M | M)$.

\paragraph{Posterior of $M$ given $D$}
The Bayesian paradigm simply places probabilities over parameters and models given the data, namely $ p( \theta_m , m | y ) $ where $ y = ( y_1 , \ldots , y_n ) $.
This has a number of decompositions.
This leads to a posterior over models, which is calculated as:
\begin{align*}
    P(M\mid D)  & = \dfrac{P(D\mid M)P(M)}{P(D)}, \\
    P(D\mid M ) & = \int_{ \Theta_M} P(D\mid \boldsymbol{\theta}_M , M ) p( \boldsymbol{\theta}_M | M ) d \boldsymbol{\theta}_M.
\end{align*}
Notice that this requires a joint prior specification $p(\boldsymbol{\theta}_M, M) = p(\boldsymbol{\theta}_M | M)p(M)$ over parameters and models. The quantity $p(M| D)$ is the marginal posterior for model complexity given the data.
There is an equivalent posterior $p(\boldsymbol{\theta}_M | D)$ for the parameters. $p(D \mid M)$ is the evidence of the data $D$ given the complexity (a.k.a. conditional likelihood). The full evidence is
\[
p( D ) = \int p( D| M ) p(M) d M.
\]
This has been used to select the amount of hyper-parameter regularization; see, for example,~\cite{mackay1992bayesian}. \cite{mackay1992bayesian} discusses Bayesian interpolation. He also discusses the correlation between training and test errors and whether they are related. Specifically, he discusses how the Bayesian paradigm deals with the trade-off between training error and model selection (a.k.a. hyperparameter regularization). The marginal Bayesian probabilities of the model (a.k.a. weight of evidence) provide a more efficient approach and computationally easier methods than traditional classical methods such as cross-validation.

\paragraph{Extending the Art of the Conversation}
\cite{bernardo2000bayesian} discusses the open and closed model framework. This relates to Peirce's theory of abduction (rather than induction). Abduction is the process of reasoning to the best explanation---given surprising observations, we form hypotheses that would explain those observations if they were true. Unlike deduction (which moves from general principles to specific conclusions) or induction (which generalizes from specific observations to general patterns), abduction involves generating new explanatory hypotheses. In the context of Bayesian model selection, abduction corresponds to introducing new models $M_{J+1}$ when existing models fail to adequately explain the observed data. Suppose another model $M_{J+1}$ is entertained by the researcher. 

We can marginalize in the other direction, over models, to obtain the posterior over the parameter of interest.  By extending the conversation, we allow ourselves to imagine the prior over parameters as including $M_{J+1}$, namely:
\[
p(\boldsymbol{\theta}) = \sum_{j=1}^{J+1} p(\boldsymbol{\theta} | M_j) p(M_j).
\]
This allows us to calculate the posterior of the new model:
\[
p(M_{J+1} | D) = \frac{p(D | M_{J+1}) p(M_{J+1})}{p(D)}.
\]
Or more simply, the relative posterior probability of $M_{J+1}$ to the set $(M_1, \ldots, M_J)$.

Then we can incorporate the new $p(M_{J+1} | y)$ into our analysis. The key point is that the other ratios of model probabilities that we calculated are unchanged!

This is simply the principle of coherence, one of the main benefits of the Bayesian paradigm. 
The evidence $p(D)$ does change, however, as we have included a new model. See also Bartlett's paradox~\cite{bartlett1957comment} and the asymptotic behavior of Bayes factors.

Observing unlikely events shouldn't necessarily change your view of model space. One philosophical question of interest is whether you can learn just from empirical observation alone and the principles of induction. \cite{diaconis1989methods} develop a theory of coincidences. Fisher, in this regard, remarks:
\begin{quote}
\emph{... for the 'one chance in a million' will undoubtedly occur with no less and no more than its appropriate frequency, however surprised we may be that it should occur to us.}
R.A. Fisher (1937, p.16)
\end{quote}

\paragraph{Model Elaboration and Inference}

Now let $ m \in \mathcal{M} $ be a family of models. 
To convey the general  flavor of the Bayesian analysis of an elaborated model suppose that we entertain a parametric model  $ f( y | \theta ) $ where $ \theta $ is an unknown parameter vector.
The family of densities
$$
\{ f ( y | \theta , m ) \; , \; m  \in \mathcal{M} \} 
$$
where $ \mathcal{M} $ is some indexing of models. Let $m \subset M$. 

Applying the usual Bayesian paradigm (a.k.a. disciplined probability accounting) to the elaborated framework, we see that inference about $ \theta $ are determined by
$$
p( \theta | y ) = \sum_{m \in M} p( \theta | m , y ) p( m | y ) 
$$
where 
\begin{align*}
 p( \theta | m , y )  & \propto p( y | \theta , m ) p( \theta | m)  \\
p( m | y ) & \propto p( y | m) p( m )  \\
{\rm where} \;  \;  p( y | m )  & = \int p( y | \theta , m ) p( \theta | m ) d \theta  
\end{align*}

For consistency across models, we take $ p( \theta | m_0, M  ) = p( \theta | M  ) $.

The posterior mean us simply a weighted average, with respect to $ p( m | y ) $, namely 
$$
\E{\theta  \mid y } = \sum_{ m \in M }  \E{\theta \mid m , y} p( m | y ) = \Ewrt{ m\mid  y}{\E{ \theta \mid m , y}}
$$
A similar decomposition holds for the variance, namely
$$
Var(\boldsymbol{\theta}\mid y) = \E{Var(\boldsymbol{\theta}\mid m,y)} +  Var\left(\E{\boldsymbol{\theta}\mid m,y}\right).
$$
The key to double decent is that for $p>n$ the Laplace approximation (a.k.a. BIC) is a poor approximation and in fact the prior $ p( \theta | m ) $ can effect the Bayes risk.
Moreover, the Bayesian folklore is that we should always fit the largest possible model as the following shows. 
\paragraph{Definition of Bayesian Double Descent}
Let $R(M) = \Ewrt{y,\boldsymbol{\theta} | M}{(\hat{\boldsymbol{\theta}}_M(y) - \boldsymbol{\theta})^2}$ be the conditional prior Bayes risk of an estimator $\hat{\boldsymbol{\theta}}_M$ derived under model $M$. Let $\hat{\boldsymbol{\theta}}_M(y) = \E{\boldsymbol{\theta} \mid y, M}$ be the optimal Bayes estimator given model $M$. The Bayesian double descent phenomenon occurs when $R(M)$ exhibits a re-descending behavior for $M > n$ (where $M$ represents model complexity and $n$ is the sample size). This re-descent is fundamentally driven by the prior specification $p(\boldsymbol{\theta}_M | M)$, which acts as regularization in the over-parameterized regime.

To be more specific, under model $M$
$$
\hat f_M = \sum_{i=1}^{M}\hat \theta_i^M\phi_i(x).
$$

Now, we introduce a different smaller model $m$, which is nested in a larger $M$ model
$$
\hat f_m = \sum_{i=1}^{m}\hat \theta_i^m\phi_i(x),
$$
and 
$$
\hat \theta_m = \E{\boldsymbol{\theta}_m \mid \boldsymbol{\theta}_{M-m} = 0, y, M}.
$$
Here $\boldsymbol{\theta}_{M-m}$ is the vector 
$\boldsymbol{\theta}_M = (\boldsymbol{\theta}_{M-m},\boldsymbol{\theta}_{m}) $.

\paragraph{Empirical Risk Bounds} It's worth noting the relationship between expected risk and empirical risk. The expected risk, $R$, is typically bounded by the empirical risk plus a term of order $1/\sqrt{N}$:
$$
R(y, f^\star) \leq \frac{1}{N} \sum_{i=1}^N R(y_i, f^\star(x_i)) + O\left(\frac{\|f\|}{\sqrt{N}}\right)
$$
The caveat is that in the case of interpolation, where the model perfectly fits the training data, the empirical risk term $\hat{R}(f)$ becomes zero. This renders such bounds uninformative for interpolators, which are central to understanding double descent. The functional norm or complexity measure is related to concepts like VC-dimension, but the key insight is that classical bounds are often unrelated to prior specifications, which are fundamental to the Bayesian perspective on double descent.

\subsection{Example: Occam's Razor}
Suppose that we only have two hypotheses $M_1$ and $M_2$. The posterior odds are given by the prior odds multiplied by the Bayes factor:
\[
    \dfrac{P(M_1\mid D)}{P(M_2\mid D)} = \underbrace{\dfrac{P(D\mid M_1)}{P(D\mid M_2)}}_{\text{Bayes Factor}} \cdot \underbrace{\dfrac{P(M_1)}{P(M_2)}}_{\text{Prior Odds}}.
\]
We can introduce Occam's razor by putting a higher prior probability $P(M_1)$ on the simpler model $M_1$. The Bayes factor itself automatically embodies Occam's razor. If $M_2$ is a more complex model, it typically spreads its prior probability mass $P(\boldsymbol{\theta}|M_2)$ over a larger parameter space. This can lead to a smaller marginal likelihood $P(D|M_2)$ unless the data strongly favor $M_2$, even if $M_2$ can fit the data perfectly. Thus, if data are fitted about equally well by both models, the simpler model will often have a higher posterior probability, especially if prior odds are equal or favor simplicity.

\paragraph{Example} To illustrate this behavior, suppose that the data are $D = (-1, 3, 7, 11)$. Now consider two hypotheses (a.k.a. models) defined by:
\begin{itemize}
    \item $M_a$: The sequence is an arithmetic progression, $y_x = n_0 + (x-1)n$, where $x=1,2,3,4$, and $n_0, n$ are integers.
    \item $M_c$: The sequence is generated by a cubic function of the form $y_x = cx^3 + dx^2 + ex + f$, where $c, d, e, f$ are fractions. (A cubic can perfectly fit 4 points).
\end{itemize}
For $M_a$, the data $D$ are uniquely generated by $n_0=-1, n=4$. Assuming $n_0, n \in \{-50, -49, \ldots, 50\}$ (a space of $101^2$ possibilities) and a uniform prior over these, we have:
\[
    P(D\mid M_a) = \dfrac{1}{101^2} \approx 0.000098.
\]
For $M_c$, a cubic polynomial $P(x) = ax^3+bx^2+cx+d_0$ can be made to pass exactly through any four points $(x_i, y_i)$. If we assume a very large discrete space of possible fractional coefficients for $c,d,e,f$ such that many combinations exist, but only one (or very few) generates $D$ exactly, then $P(D|M_c)$ would be (number of fitting parameters) / (total parameter combinations). If the parameter space for $M_c$ is much larger than for $M_a$, $P(D|M_c)$ could be smaller than $P(D|M_a)$, even though $M_c$ can also fit the data.
(Note: The original example for $M_c$ with specific fractions and counts was unclear and potentially flawed; this is a conceptual rewording).
Note that we made an implicit assumption that the prior probability on the parameters is uniform within the defined ranges. This assumption is subjective and is not always the appropriate one. For example, if we know that the sequence is generated by a smooth function, we can put a prior on the parameters that is more likely to generate a smooth curve.

Having established how Bayesian methods naturally incorporate Occam's razor through the marginal likelihood, we now turn to the main contribution of this paper: understanding double descent from a Bayesian perspective.

\subsection{BIC \texorpdfstring{$m < n$}{m < n}}
For $ p<n $, Laplace's approximation provides a simple yet powerful  \cite{lindley1961use} approximation of how dimensionality is weighted in the Bayesian paradigm.  In its simplest form we recover the Schwarz BIC criterion.  Namely 
$$ m_k = \int_{ \Re^k }  f( y | \theta_k ) p( \theta_k ) d \theta  \approx n^{- k/2} f( y | \hat{\theta}_k ) $$ 
Notably, the approximation
doesn't depend on the prior $p(\theta) $. On a log-scale, we find the model that maximises
$$
\log f ( y | \hat{\theta}_k ) - \frac{k}{2} \log n 
$$
The $ -(k/2) \log n $ term acts as an Occam's razor, favoring models of lower dimensions, all else being equal. 

The Laplace approximation gives the closer approximation 
$$
\hat{m}_j =  (2 \pi)^{d_j/2} \mathcal{L}_j  ( \hat{\theta}_j )  | \det I  ( \hat{\theta}_j )  |^{\frac{1}{2}}
$$
Hence BIC is related to log-posterior approximation. Again there is an Occam term. DIC provides a further approximation for exponential families. 

The Bayesian Information Criterion (BIC) is a model selection criterion that penalizes the complexity of the model. It is derived from a Bayesian approach. The BIC is defined as:
\[
    \mathrm{BIC} =  \log P(D\mid \hat{\boldsymbol{\theta}}_m, m) - \frac{m}{2} \log n.
\]
Here $\hat{\boldsymbol{\theta}}_k$ is the maximum likelihood estimate of the $k$ parameters in model $M_k$, and $n$ is the sample size. As such, there is a penalty $-\frac{k}{2} \log n$ for increasing the dimensionality $k$ of the model under consideration.

In posterior model probability space, we  then have the approximation
$$
p( M | D ) \approx \exp \left ( - \frac{1}{2} BIC \right )  
$$
Convergence in model space to the true model can happen very quickly--at an exponential rate, see Appendix A for details.

To further understand  BIC, we need to understand the Bayesian approach to model selection. Specifically, the marginal likelihood of the data under model $M_k$ (denoted $M$ for simplicity here) is approximated using Laplace's method:
\[
    P(D\mid M) = \int P(D\mid \boldsymbol{\theta},M)P(\boldsymbol{\theta}\mid M)d\boldsymbol{\theta} \approx P(D\mid \hat{\boldsymbol{\theta}},M)P(\hat{\boldsymbol{\theta}}\mid M) (2 \pi)^{m/2} |\det(\mathbf{H}(\hat{\boldsymbol{\theta}}))|^{-\frac{1}{2}}.
\]
Here $\hat{\boldsymbol{\theta}}$ is the posterior mode (MAP estimate), and $\mathbf{H}(\hat{\boldsymbol{\theta}})$ is the negative Hessian of the log-posterior at the mode.
Taking the logarithm, and assuming $P(\hat{\boldsymbol{\theta}}|M)$ and Hessian terms are $O_p(1)$ or scale appropriately with $n$, we get:
\[
    \log P(D\mid M) \approx \log P(D\mid \hat{\boldsymbol{\theta}}_{\text{MLE}},M) - \dfrac{m}{2}\log n,
\]
which is proportional to the BIC. (Note: The exact definition and derivation of BIC can vary slightly, but this captures the essence).
The BIC approximation shows how the Bayesian approach naturally penalizes model complexity through the dimensionality term $-\frac{k}{2}\log n$.

The Bayesian approach averages over the posterior distribution of models given data. 
Suppose that we have a finite list of models $M \in \{M_1, \ldots, M_J\}$. Then we can calculate the posterior over models as:
\[
p(M_j | y) = \frac{p(y | M_j) p(M_j)}{\sum_{i=1}^J p(y | M_i) p(M_i)}, \quad {\rm where} \; p(y | M_j) = \int L_j(\boldsymbol{\theta}_j|y) p(\boldsymbol{\theta}_j | M_j) d\boldsymbol{\theta}_j.
\]
Laplace's approximation provides a simple~\cite{lindley1961use} illustration of how dimensionality is weighted in the Bayesian paradigm. Hence, BIC is related to a log-posterior approximation. Hence, if prior model probabilities $P(M_j)$ are uniform, then $P(M_j\mid D) \propto P(D \mid M_j) \approx \exp(\mathrm{BIC}_j)$.

In a more general case, the evidence (a.k.a. marginal likelihood) for hypotheses (a.k.a. models) $M_i$ is calculated as follows:
\[
    P(D\mid M_i) = \int P(D\mid \boldsymbol{\theta}, M_i)P(\boldsymbol{\theta}\mid M_i)d\boldsymbol{\theta}.
\]
Laplace approximation, in the one-dimensional case ($k=1$), yields:
\[
    P(D\mid M_i) \approx P(D\mid \hat{\boldsymbol{\theta}}, M_i)P(\hat{\boldsymbol{\theta}}\mid M_i)\sqrt{2\pi}\sigma_{\text{post}}.
\]
Here $\hat{\boldsymbol{\theta}}$ is the maximum \emph{a posteriori} (MAP) estimate of the parameter and $\sigma_{\text{post}} = (-H(\hat{\boldsymbol{\theta}}))^{-1/2}$ where $H(\hat{\boldsymbol{\theta}})$ is the second derivative of the log-posterior at $\hat{\boldsymbol{\theta}}$.

Generally, in the $k$-dimensional case, we have:
\[
    P(D\mid M_i) \approx P(D\mid \hat{\boldsymbol{\theta}}, M_i)P(\hat{\boldsymbol{\theta}}\mid M_i) (2\pi)^{k/2} |\det(-\mathbf{H}(\hat{\boldsymbol{\theta}}))|^{-\frac{1}{2}}.
\]
Here $\mathbf{H}(\hat{\boldsymbol{\theta}}) = \nabla^2\log (P(D\mid \hat{\boldsymbol{\theta}}, M_i)P(\hat{\boldsymbol{\theta}}\mid M_i))$ is the Hessian of the log-posterior function evaluated at the mode $\hat{\boldsymbol{\theta}}$. As the amount of data collected increases, this Gaussian approximation is expected to become increasingly accurate.

\cite{mackay1992bayesian} proposes the NIC criterion for selection of neural networks.

The key to double descent is that for $p > n$ the Laplace approximation (a.k.a. BIC) is a poor approximation and in fact the prior $p(\boldsymbol{\theta} | m)$ can affect the Bayes risk. Moreover, the Bayesian folklore is that we should always fit the largest possible model as the following shows.

\section{Double Descent and Global-Local Shrinkage}\label{sec:grr-shrinkage}

Our approach is to connect double descent with the Bayesian principles of global-local shrinkage. We illustrate our methodology in a number of settings. 

\subsection{Over-parameterisation, \texorpdfstring{$m>n$}{m > n}} 

Bayesian folklore is that one should fit  a model as large as an elephant \citep{savage1956foundations} (although an elephant can be parameterised by 4 parameters!).

For example, \cite{deaton1980empirical} and \cite{young1977bayesian} propose priors for polynomial regression with probability of polynomial degree decaying. \cite{young1977bayesian} argues that "the best approach to prediction using polynomials is to fit the largest possible degree commensurate with our computing and statistical skills". The theorem below formalizes this statement. 

\begin{theorem}[Model Nesting and Computational Equivalence]
To compute $P(m\mid y)$ using the Dickey-Savage density ratio, we also need to calculate $\hat f_m(x)  = E \left[f(x,\theta_m \mid \theta_{M-m} = 0,y,M)\right]$.
\end{theorem}
Consider the functional regression models
$$
y = \sum_{m=1}^M  \theta_m^T f_m ( x )  + \epsilon 
$$
where $ y = ( y_1 , \ldots , y_n ) $ and $ F_m(x) = (  f_m ( x_1 ) , \ldots , f_m ( x_n ) ) $ is a stacked set of predictors.

Then under the consistency of likelihoods and priors,  namely 
$$ 
p( \theta_m | m ) = p( \theta_m | \theta_{m+1:M} =0 , y )  \; {\rm and} \; p( y | \theta_m , m ) = p( y | \theta_m ,  \theta_{m+1:M} =0  )
$$ 
where $ \theta_m = ( \theta_1 , \ldots , \theta_m ) $ and $ \theta_{m+1:M} = ( \theta_{m+1} , \ldots , \theta_M  ) $.
Then we can calculate our functional estimate for
sub-model $m$ from the over-parameterised full model $M$, via 
\begin{align*}
\hat{f}_m ( x) & = \E{\hat{f}_m (x) | \boldsymbol{\theta}_{m+1:M}=0 , M , y}
\end{align*} 

Hence, you can calculate probabilities under the nested simpler model $m$ from the full model $M$. This is simply a question of computational power.

\begin{proof}
This is the Dickey-Savage density ratio: write the parameter vector as $\boldsymbol{\theta}_M = (\boldsymbol{\theta}_m, \boldsymbol{\theta}_{M-m})$ and assume the priors are "consistent" across specifications. Then we have the result:
\[
\frac{p(y|m)}{p(y|M)} = \frac{p(\boldsymbol{\theta}_{m+1:M}=0 | y, M)}{p(\boldsymbol{\theta}_{m+1:M}=0 | M)}
\]
You just need the posterior ordinates for $\boldsymbol{\theta}_{M-m} = 0$. Also probabilities calculated under the big model $M$.

The above Bayesian accounting naturally leads to the Dickey-Savage density ratio, which provides a powerful tool for comparing nested models within this framework.

We are assuming that the likelihood for model $m$ is nested in $M$ in the sense that $p(y | \boldsymbol{\theta}_m) = p(y | \boldsymbol{\theta}_m, \boldsymbol{\theta}_{M-m} = 0)$ and equivalently for the priors:
\[
p(\boldsymbol{\theta}_{M-m} = 0 | y) = \int p(\boldsymbol{\theta}_{M-m} = 0, \boldsymbol{\theta}_m | y) d\boldsymbol{\theta}_m = \int \frac{p(y | \boldsymbol{\theta}_{M-m} = 0, \boldsymbol{\theta}_m) p(\boldsymbol{\theta}_{M-m} = 0, \boldsymbol{\theta}_m)}{p(y)} d\boldsymbol{\theta}_m
\]
Assuming conditional independence in the prior $p(\boldsymbol{\theta}_{M-m} = 0, \boldsymbol{\theta}_m) = p(\boldsymbol{\theta}_{M-m} = 0 | \boldsymbol{\theta}_m) p(\boldsymbol{\theta}_m)$ under $M$, we have:
\[
p(\boldsymbol{\theta}_{M-m} = 0 | D) = \int p(\boldsymbol{\theta} = 0, \boldsymbol{\phi} | D) d\boldsymbol{\phi} = \int \frac{p(y | \boldsymbol{\theta} = 0, \boldsymbol{\theta}_m) p(\boldsymbol{\theta}_m) d\boldsymbol{\theta}_m}{p(y)} p(\boldsymbol{\theta}_{M-m} = 0)
\]
Hence, as $p(y | \boldsymbol{\theta}_{M-m} = 0) = \int p(y | \boldsymbol{\theta}_{M-m} = 0, \boldsymbol{\theta}_m) p(\boldsymbol{\theta}_m) d\boldsymbol{\theta}_m$, we have:
\[
\frac{p(\boldsymbol{\theta}_{M-m} = 0 | y,M)}{p(\boldsymbol{\theta}_{M-m} = 0)} = \frac{p(y | \boldsymbol{\theta}_{M-m} = 0)}{p(y)} = \frac{p(y|m)}{p(y|M)}
\]
All calculated under $M$ on the left-hand side!

Hence, we can calculate the posterior over sub-models $m$, purely from $p(\boldsymbol{\theta}_M | y, M)$ and hence:
\[
p(m|y) = \frac{p(\boldsymbol{\theta}_{M-m} = 0 | y,M)}{p(\boldsymbol{\theta}_{M-m} = 0)} p(m)
\]
\end{proof}

\paragraph{Regression with $p<n$} Consider a regression model with $n\times p$  design matrix $X$. Consider case when $p<n$. We can always find orthogonal matrices $ P $  and $Q$ such that
$$
P X^T X Q = \Lambda^2 \; \; {\rm and} \; \;  P X Q^T = diag ( \Lambda \;  | \; 0 )  = D 
$$
Let $ Z = P y $ and $ \gamma = Q \beta $ and  $ \nu = P \epsilon $. Then we have the model
$$
Z = D \gamma + \nu 
$$
We have hierarchical  model, 
\begin{align*}
z_i  | \gamma_i , \sigma^2 & \sim N ( \lambda_i \gamma_i , \sigma^2 ) , 1 \leq i \leq p  \; {\rm  and} \\
 z_i  | \sigma^2 &  \sim N( 0 , \sigma^2 ) \; {\rm for} \;   i = p+1 , \ldots , n  \\
\gamma_i  & \sim N( 0 , \sigma_{ \gamma_i }^2  )
 \end{align*}
Give, the OLS estimate $\beta^\star$, the estimator $ \gamma^\star $ of $ \gamma $ is given by a shrinkage estimator
$$
\gamma^\star = Q \beta^\star = ( \Lambda^2 +  K )  D^T Z = \frac{ \lambda_i^2 }{ \lambda_i^2 + k_i } \hat{\gamma}_i 
$$
where $ \hat{\gamma}_i = z_i / \lambda_i $ is the least squares estimate. 

Here $ K = diag ( k_1 , \ldots , k_p ) $ is a vector of generalised ridge hyper-parameters.  The local scales of a global-local shrinkage model. 

The optimal choice of $ k_i $ is given by optimizing the marginal likelihood $ z_i | k_i , \sigma^2 $ which is given by 
$$
p( z_i | k_i ) \sim N \left ( 0 , \sigma^2 \left ( 1 + \frac{\lambda_i^2}{k_i} \right )  \right )
$$
Let $ \phi_i^2 = \sigma^2 \left ( 1 + \frac{\lambda_i^2}{k_i}  \right ) $. Then by 1-1 transformation property of MLE we have
$$
\hat{k}_i =  \frac{ \lambda_i^2 \sigma^2 }{  \hat{\phi}_i^2 - \sigma^2 }= \frac{ \lambda_i^2 \sigma^2}{ z_i^2 - \sigma^2 } 
$$
As  $ k_i > 0 $. The GRR estimator (given $ \sigma^2$) is a variable selection estimator of the form
\begin{align*}
\hat{k}_i & = \frac{ \lambda_i^2 \sigma^2}{ z_i^2 - \sigma^2 }  \; {\rm for} \; z_i^2 > \sigma^2 \\
 & = \infty   \; {\rm for} \; z_i^2 < \sigma^2 
\end{align*}
where $ k_i = \infty $ corresponds to variable selection.

\subsection{Interpolation, \texorpdfstring{$m=n$}{m = n}}

\paragraph{Gaussian Means} Suppose that we have the infinite Gaussian means problem:
\begin{align*}
y_i |  \boldsymbol{\theta}_i &  \sim N(\boldsymbol{\theta}_i, \sigma^2), \\
\boldsymbol{\theta}_i &    \sim N(0, \tau^2).
\end{align*}
This is the well-known variance components ANOVA model. There are many subtle issues in the joint estimation of $(\sigma^2, \tau^2)$; see~\cite{hill1965inference} and~\cite{tiao1965bayesian}. \cite{polson2012local} provide the global-local shrinkage framework to analyze these models from a fully Bayesian perspective.

There is a long history of studied estimation procedures in these models. For example, \cite{neyman1948consistent} shows that the MLE is inconsistent for estimating $\sigma^2$. This is due to the fact that one has to marginalize out the infinite-dimensional nuisance parameter $\boldsymbol{\theta} = (\boldsymbol{\theta}_1, \boldsymbol{\theta}_2, \ldots)$. \cite{hill1965inference} shows that the classical approach can lead to negative variance estimates. This is just a symptom of the fact that there is a spike at the origin in the marginal likelihood for the variance. Hence, the Bayesian paradigm and careful specification of priors are required. Inverse Gamma priors shrink \emph{away} from the origin and hence do not support the prior assumption of sparsity. \cite{polson2012local} discuss the use of horseshoe (a.k.a. half-Cauchy scale priors) on the variance components and discuss the analytical behavior of the marginal likelihood.

The heterogeneous variance case is also of interest:
\begin{align*}
y_i |  \boldsymbol{\theta}_i &  \sim N(\boldsymbol{\theta}_i, \sigma^2), \\
\boldsymbol{\theta}_i &    \sim N(0, \tau_i^2).
\end{align*}
Now we wish to estimate a sequence of prior hyperparameters (a.k.a. regularization parameters) $\mathcal{T} = (\tau_1^2, \tau_2^2, \ldots)$.
This will be most useful for the hierarchical model case. 

\cite{xu2007some} shows that we need to be careful with the Bayes risk when all the $\tau_i$ are zero, as it is infinite. We are expecting non-zero sparsity.
A simple estimator for $\tau_k^2 + \sigma^2$ is $y_k^2$ (if $\E{\boldsymbol{\theta}_k}=0$). We can achieve a better estimator by using empirical Bayes and an adaptive schedule. This requires ordering of the parameters.
The Bayes risk can be calculated as follows. The choice of hyperparameter schedule $\{\tau_i^2\}$ is crucial: if the true $\tau_k^2$ decay at a certain rate but this rate is misspecified in the prior, it can lead to suboptimal risk behavior and affect the double descent phenomenon.

Interpolation is equivalent to the normal means problems: $n$ signals and $ n$ means to be inferred. 
Consider the sparse normal means model $y_i = \theta_i + \epsilon_i$ where
$$
\theta_i | \kappa_i  \sim N \left ( 0 , \frac{1- \kappa_i}{ \kappa_i} \right ) 
$$
This corresponds to the class of horseshoe priors. Bhadra et al. recommend these as default priors for nonlinear functionals.
In a  regression setting
\[
\hat \alpha_i = \alpha_i + \epsilon_i,
\]
Then the optimal Bayes estimator is 
\[
\alpha^*_i = w_i\hat \alpha_i = (1-\kappa_i)\hat \alpha_i.
\]
Assuming the prior $\alpha \sim N(0,V)$, $V = \text{diag}(v_1,\ldots,v_p)$, then 
\[
p(\alpha_i\mid w_i) = \sqrt{\dfrac{\lambda_i(1-w_i)}{2\pi w_i}}\exp\left(-\dfrac{\lambda_i(1-w_i)}{2w_i}\alpha_i^2\right) \sim N\left(0,\dfrac{w_i}{\lambda_i(1-w_i)}\right).
\]
The log-likelihood is
\[
\log p(D\mid \alpha, \sigma^2) = c - \dfrac{n}{2}\log \sigma^2 - \dfrac{1}{2\sigma^2}||Y-X\alpha||_2^2 =  c - \dfrac{n}{2}\log \sigma^2 + \dfrac{1}{2} \sum_{i=1}^{p}\left\{\log(1-w_i) + \dfrac{w_i\lambda_i\hat \alpha_i^2}{\sigma^2}\right\}.
\]
For regularization priors, $w_i \sim \text{Beta}(a,b)$. Note that when $a=b=1/2$ this leads to the Horseshoe model. Let $t_i = \sqrt{\lambda_i}\hat \alpha_i/\sigma$. Many authors use deterministic $W(t)$, e.g., Empirical Bayes or Type II maximum likelihood \cite{polson2012local}.

Finally, much theoretical research on double descent has focused on the min-$L^2$-norm estimator with Moore-Penrose pseudo-inverse of $X^T X$, which is the solution to the optimization problem (see~\cite{hastie2022surprises, bach2024highdimensional}): 
$$
\hat{\beta} = ( X^T X)^+ X^T y  = \arg \min \{ \vnorm{b}_2 : b \; {\rm minimizes} \; \vnorm{y - X b}^2_2 \} 
$$
with stochastic regressors.
The risk goes to infinity at the interpolation point. The Belkin plot simply demonstrates that a diffuse Lebesgue prior is a poor choice in high dimensions, see also~\cite{efron1973steins}.

\section{Bayesian Double Descent and NN Regression}\label{sec:bayesian-double-descent}

Consider the functional model $y_i = f(x_i) + \epsilon_i$, for a training set $D = \{(x_i,y_i)\}_{i=1}^n$, the posterior Bayes risk for an estimator $\hat{f}$ of $f$ is:
\[
    R(\hat{f} | D) = \int L(\hat{f},f)P(f\mid D)df.
\] 
For the model complexity problem, we can also consider the prior Bayes risk (or integrated risk), which averages over data $y$, true parameters $\boldsymbol{\theta}$, and models $M$. For squared error loss:
\[
    R_{\text{prior}}(\hat{\boldsymbol{\theta}}) = \Ewrt{y,\boldsymbol{\theta},M}{(\hat{\boldsymbol{\theta}}(y) - \boldsymbol{\theta})^2} = \Ewrt{y}{\Ewrt{M\mid y}{\Ewrt{\boldsymbol{\theta}\mid M,y}{(\hat{\boldsymbol{\theta}}(y) -\boldsymbol{\theta})^2}}}.
\]
Notice that the optimal estimator (Bayes estimator) under squared error loss is the posterior mean $\hat{\boldsymbol{\theta}}_{\text{Bayes}}(y) = \E{\boldsymbol{\theta}\mid y}$. By the law of iterated expectations:
\[
    \hat{\boldsymbol{\theta}}_{\text{Bayes}}(y) = \E{\boldsymbol{\theta}\mid y} = \Ewrt{M\mid y}{\E{\boldsymbol{\theta}\mid M,y}}.
\]
Now, we can see a potential ``flaw'' in interpreting the typical double descent plot (Figure~\ref{fig:double-descent}) from a purely Bayesian perspective if it only shows $R(\hat{\boldsymbol{\theta}}(M),\boldsymbol{\theta})$ (a frequentist risk for an estimator derived from model $M$). A fully Bayesian approach would average over $p(M\mid y)$. The overall posterior Bayes risk for $\hat{\boldsymbol{\theta}}_{\text{Bayes}}(y)$ can be written as:
\[
    R(\hat{\boldsymbol{\theta}}_{\text{Bayes}} | D) = \Ewrt{M\mid D}{R_M(\hat{\boldsymbol{\theta}}_M | D)},
\]
where $\hat{\boldsymbol{\theta}}_M = \E{\boldsymbol{\theta} \mid D, M}$ and $R_M(\hat{\boldsymbol{\theta}}_M | D) = \Ewrt{\boldsymbol{\theta} \mid D,M}{(\hat{\boldsymbol{\theta}}_M - \boldsymbol{\theta})^2}$.

\paragraph{The $p>n$ Case} In high-dimensional settings where $p > n$, Bhattacharya and Mallick use the Sherman-Woodbury formula. Everything can be reduced to an $n \times n$ interpolation case. The Bayes risk goes to infinity when the local prior variances are equal to $\tau^2$ and this tends to infinity.
$$
( X^T X + D^{-1} )^{-1} = D - D X^T ( X D X^T  + I_n )^{-1} X D 
$$
The classical literature places much emphasis on the minimum norm $L^2$-estimator where $\hat{\beta} = (X^T X)^+ X^T y$ where $(X^T X)^+$ is the Moore-Penrose inverse. They consider the $X$-random case.

$D$ is the matrix of local shrinkage factors.

Breiman (1995) shows that the Non-Negative Garotte is equivalent to the positive part James-Stein estimator. He shows that it outperforms subset regression in terms of predictive risk. See also Frank and Friedman (1993). He considers $X$-controlled and $X$-random input cases. Breiman and Spector (1992) consider the use of cross-validation and prediction error.

Goldstein and Smith \citep{goldstein1974ridgetype} provide conditions under which this class of estimators dominates the least-squares solution in componentwise mean squared error (MSE). Specifically, if the smallest singular value $d_{\min}^2$ exceeds a threshold determined by the signal-to-noise ratio, then:
\[
\text{MSE}(\boldsymbol{\alpha}_i^*) < \text{MSE}(\hat{\boldsymbol{\alpha}}_i)
\]
for all $i$. This contrasts with James--Stein estimators, which guarantee overall MSE reduction but not improvement for each parameter individually.

While the James--Stein estimator also shrinks the OLS coefficients toward zero, it does so uniformly across all coordinates through a global multiplier that depends on the total squared norm $\|\hat{\boldsymbol{\alpha}}\|^2$. In contrast, the GRR estimator performs shrinkage independently along each coordinate, based on local information---specifically, the corresponding singular value $d_i$ and penalty $k_i$. This local, componentwise shrinkage anticipates the structure of modern global-local shrinkage priors, where each parameter receives its own adaptive regularization governed by local scale terms. In both settings, the goal is to suppress estimation error more aggressively in directions of low signal-to-noise, but GRR achieves this in a fully decoupled fashion, which enables componentwise risk improvements that the James--Stein estimator cannot guarantee.

\paragraph{MSE and Componentwise Improvements} 
Walker and Page provide a comprehensive summary of the MSE calculations. The ridge estimates are given by
$$
\beta^\star = ( X^T X + k I_p )^{-1} X^T y \; {\rm and} \; \gamma^\star = ( D^2 + k I_p )^{-1} D^T Z
$$
In the orthogonal components representation:
$$
\gamma_i^\star = \frac{ d_i z_i }{ d_i^2 + k } = \frac{ d_i^2 }{ d_i^2 + k } \hat{\gamma}_i .
$$
The original ridge estimates (cf. Efron and Morris) use $k = p \hat{\sigma}^2 / \hat{\beta}^T \hat{\beta}$ and $1/F$ Lawless. Dempster et al. simulation study. 

Goldstein and Smith show the remarkable result that as long as the smallest eigenvalue is large enough, the Bayes GRR estimator dominates the MLE on ALL components. This is a stronger result than the James-Stein estimator, which only provides overall risk improvement.

On an intuitive level, the James-Stein form is inappropriate in the sense that it implicitly takes less account of the loss in precisely those directions where estimation is most inaccurate (i.e., those corresponding to small eigenvalues). In fact, an estimator of the form
\[
\gamma^*_i = \left(1-c/\sum_{j=1}^{p}z_j^2\right)\hat \gamma_i, ~ (i = 1,2,\ldots,p)
\]
has smaller mean-square error than least squares if and only if the smallest eigenvalue is greater than $(c+4)/2\sum\lambda^{-2}_i$. 

The James-Stein results refer throughout to average mean-square error. Nothing can be said concerning improved performance in the individual component parameters. The GRR approach, on the other hand, assures us of the potential for componentwise improvement. This applies also to the original parameters, where we define
\[
\beta^* = Q^T\gamma^*, \quad \gamma^*_i = c(\lambda_i, k) z_i \quad (i = 1, 2, \ldots, p).
\]

It is sufficient to show that $(d/dk)\E{\beta_i^* - \beta_i}^2\mid k=0 \le 0$.

Furthermore, one can evaluate predictive performance using the expected predictive mean squared error (PMSE), which for shrinkage weights $\kappa_i = \kappa(d_i, k_i)$ becomes:
\begin{equation}
\text{PMSE}(\tilde{\boldsymbol{\beta}}) = \sum_{i=1}^p \int \left\{ \kappa_i z_i - \mu_i \right\}^2 dF_{\mu_i}(z_i),
\end{equation}
where $z_i | \mu_i \sim N(\mu_i, 1)$, and $\mu_i$ represents the true canonical coefficient.

We can compare GRR estimators in terms of their predictive mean squared error (PMSE) as discussed by \cite{lawless2011statistical}:
\[
\text{PMSE}(\tilde{\beta}) = \|\tilde{Y} - Y\|^2 = \|X(\tilde{\beta} - \beta)\|^2 = \sum_{i=1}^p \lambda_i \mathbb{E}\left\{ (\tilde{\alpha}_i - \alpha_i)^2 \right\} / \sigma^2,
\]
where, assuming normality and taking $\sigma$ as known (or fixed), we find that
\begin{equation}
\text{PMSE}(\tilde{\beta}) = \sum_{i=1}^p \int_{-\infty}^{\infty} \left\{ W(z_i) z_i - \mu_i \right\}^2 dF_{\mu_i}(z_i),
\end{equation}
where $W(z_i)$ is the shrinkage weight, $z_i$ is the observed statistic, $\mu_i$ is the true mean, and $F_{\mu_i}(z_i)$ is the distribution function of $z_i\sim N(\mu,1)$.

This provides an alternative to isotonic regression approaches.

Thus, generalized ridge regression offers a flexible, interpretable alternative to uniform shrinkage methods, with clear Bayesian and risk-based justification. This framework also accommodates more flexible priors on $k_i$, such as inverse-gamma or half-Cauchy, yielding global-local priors with adaptive shrinkage \citep{carvalho2010horseshoe, polson2012local}.

\section{Applications}

Now, we demonstrate how hierarchical Global-Local models can provide a principled approach to handling the double descent phenomenon. Unlike the frequentist polynomial regression example above, the Bayesian framework naturally incorporates uncertainty about model parameters through prior distributions, which can lead to automatic regularization effects even in over-parameterized settings.

This hierarchical structure allows the model to adaptively determine which polynomial coefficients should be shrunk toward zero. When $d > n$, the Bayesian posterior automatically balances between fitting the data and maintaining reasonable parameter values through the prior structure.  The key advantage of this Bayesian approach is that it provides a coherent probabilistic framework for reasoning about uncertainty in over-parameterized models. While the frequentist approach relies on implicit regularization through the minimum-norm solution, the Bayesian approach makes the regularization explicit through the prior structure.

We now examine two classical Bayesian approaches that provide principled methods for polynomial model selection and regularization. Young~\cite{young1977bayesian} introduces a hierarchical prior structure that automatically penalizes higher-order polynomial coefficients, effectively implementing Occam's razor through the prior specification. Deaton~\cite{deaton1980empirical} extends this framework by allowing for different shrinkage levels across polynomial coefficients while maintaining an ordering constraint that reflects the intuition that higher-order terms should be more heavily regularized. Both approaches demonstrate how Bayesian methodology can naturally handle the bias-variance tradeoff in polynomial regression through careful prior specification, providing an alternative to the implicit regularization observed in over-parameterized frequentist models.

\subsection{Bayesian Polynomial Interpolation}\label{sec:bayesian-interpolation}

We now study the double descent phenomenon in a model for Bayesian interpolation. See~\cite{mallick1998bayesian} and~\cite{deaton1980empirical} for a discussion of polynomial regression and related Bayesian approaches. Formally, consider a statistical model of the form:
\[
    y = f(x) + \epsilon, \quad \mathrm{var}(\epsilon_i) = \sigma^2.
\]
Polynomial regression then uses a parametric model $f(x,\boldsymbol{\theta})$, where $\boldsymbol{\theta}$ is the parameter vector:
\[
    f(x,\boldsymbol{\theta}) = \sum_{j=1}^d\theta_j\phi_j(x),
\]
where $\phi_j$ are basis functions, for example, orthogonal polynomials. A special case of this model is non-parametric regression, when $q$ goes to infinity.

\paragraph{Example:} First, we  demonstrate the double descent by fitting a polynomial regression using an ordinary least squares.  We generate data from a noisy $\sin(x)$ function
\[
    y_i = \sin(5x_i) + \epsilon, ~ \epsilon \sim N(0,0.3^2), ~ n=1,\ldots,20.
\]
The choice of a small sample size $n=20$ is crucial for observing double descent, as it creates a regime where the number of model parameters can substantially exceed the number of observations.

We fit polynomial models of varying degrees $d = 1, 2, \ldots, 50$ using Legendre polynomial basis functions. Legendre polynomials provide a numerically stable orthogonal basis that helps avoid the numerical instabilities associated with standard monomial bases in high-degree polynomial fitting. For each degree $d$, we estimate the coefficients using the Moore-Penrose pseudoinverse, which provides the minimum-norm solution when the system is overdetermined (i.e., when $d > n$).

\begin{figure}
    \centering
    \includegraphics[width=0.6\textwidth]{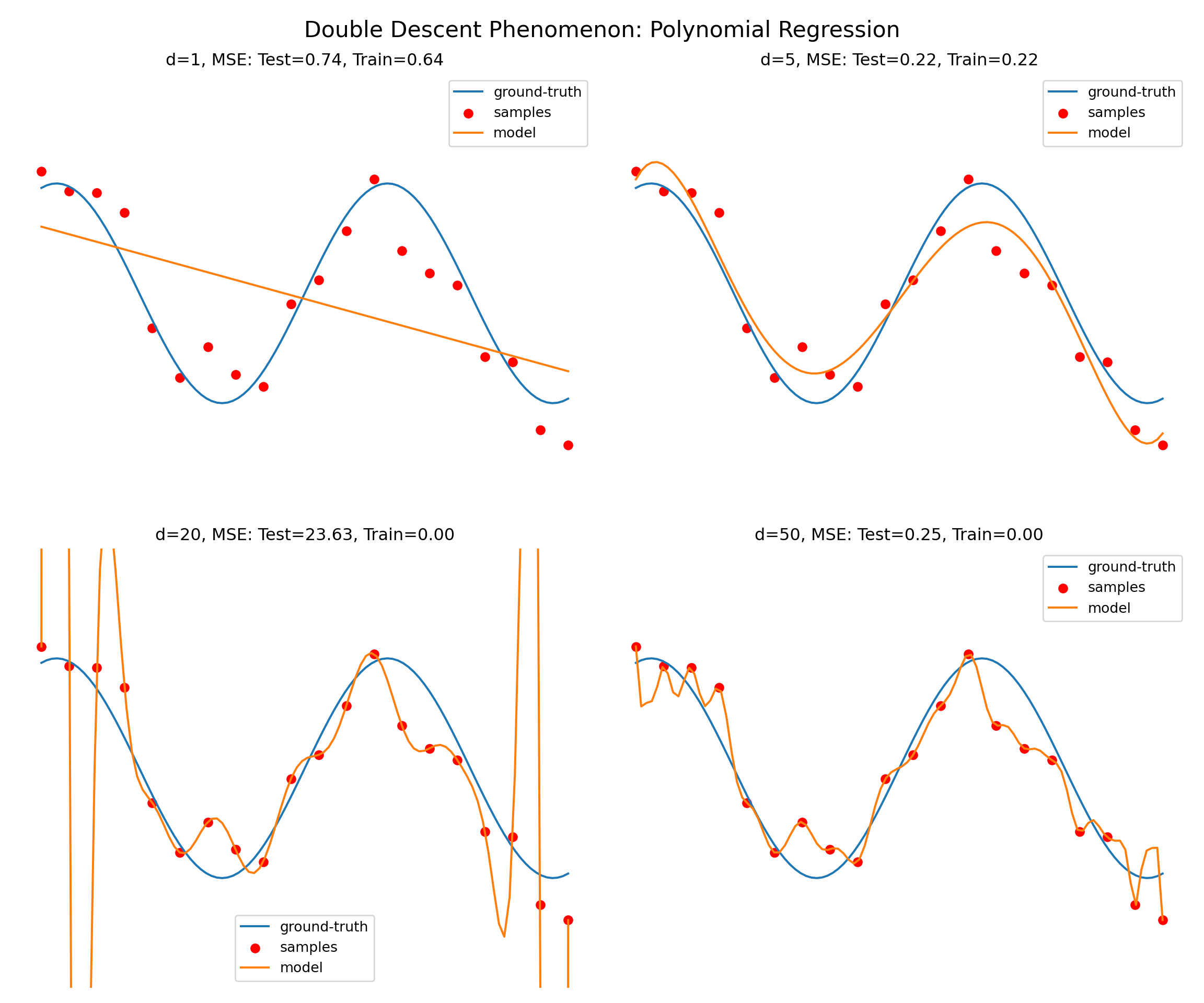}
    \caption{Double Descent Phenomenon: Polynomial Regression with Different Degrees}
    \label{fig-double-descent-grid}
\end{figure}

Now, let's plot the MSE curve. We will plot the test error (blue line) and the training error (red line) for different polynomial degrees from 1 to 50.
\begin{figure}
    \centering
    \includegraphics[width=0.6\textwidth]{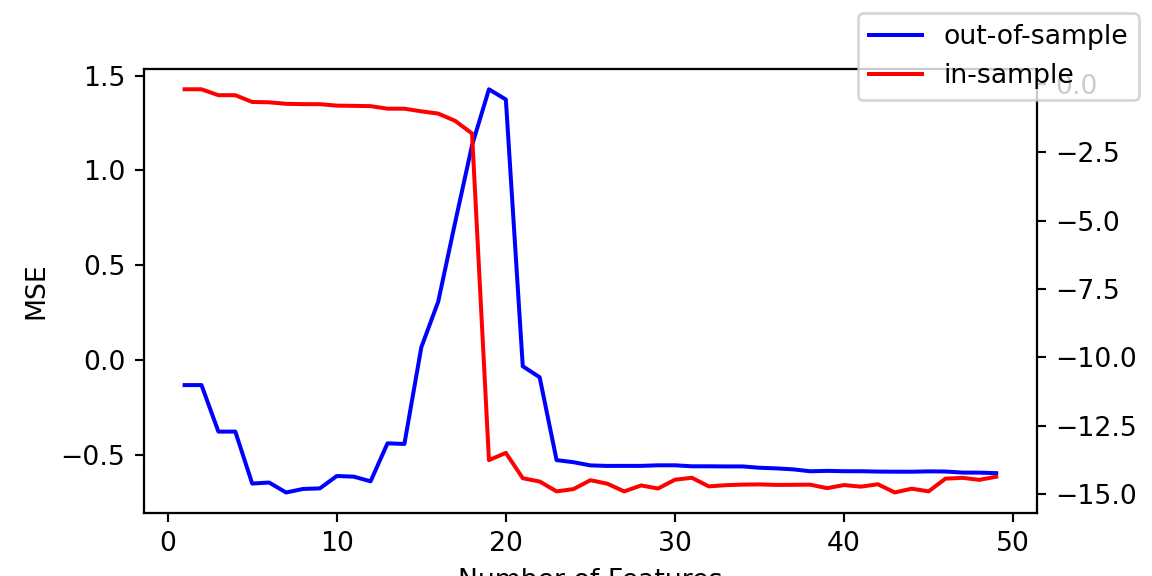}
    \caption{Bias-Variance Trade-off: Training and Test MSE vs Model Complexity}
    \label{fig-mse-curve}
\end{figure}

Figure \ref{fig-double-descent-grid} illustrates how model behavior changes dramatically across different polynomial degrees. The four panels show representative cases that capture the key phases of the double descent phenomenon. For degree 1 (underparameterized), the linear model is too simple to capture the oscillatory nature of the underlying sine function. As a result, it exhibits high bias and provides a poor fit to both the training and test data. At degree 5, we reach the classical optimum. Here, the model complexity is well-matched to the underlying function, striking a balance between bias and variance. The model is sufficiently flexible to capture the main features of the sine function without succumbing to severe overfitting. 

When the degree increases to 20, the model reaches the interpolation threshold. At this point, the number of parameters matches the number of training observations, allowing the model to perfectly interpolate the training data. However, this perfect fit comes at a cost: the fitted curve displays wild oscillations between data points, resulting in poor generalization to new data. Finally, at degree 50, the model is highly over-parameterized, with far more parameters than observations. Surprisingly, in this regime, the model achieves better test performance than at the interpolation threshold. This improvement in generalization, despite the model's capacity to memorize the training data, is a hallmark of the double descent effect.

The key insight from Figure \ref{fig-mse-curve} is the characteristic double descent shape in the test error (blue line). The curve exhibits three distinct phases:

\begin{enumerate}
    \item Classical Regime: For low degrees ($d < 5$), increasing model complexity reduces both bias and test error, following the traditional understanding of the bias-variance tradeoff.
    \item Interpolation Crisis: Around the interpolation threshold ($d \approx n = 20$), test error peaks dramatically as the model begins to perfectly fit the training data while generalizing poorly.
    \item Over-parameterized Regime: For very high degrees ($d > 30$), test error decreases again, demonstrating that extreme over-parameterization can lead to improved generalization despite the model's ability to memorize the training data.
\end{enumerate}

This behavior challenges the conventional wisdom that more parameters necessarily lead to worse generalization. The double descent phenomenon arises from the implicit regularization effects of minimum-norm solutions in over-parameterized settings. When $d > n$, the pseudoinverse solution corresponds to the minimum $\ell_2$-norm coefficients among all possible interpolating solutions. This implicit bias toward simpler functions can lead to surprisingly good generalization properties.

While this example uses polynomial regression for clarity, the double descent phenomenon has been observed across a wide range of modern machine learning models, including deep neural networks, random forests, and kernel methods. The implications for practice are significant. Given that model selection is time consuming and computationally expensive, this example shows, that instead of spending time to do model selection to find the "sweet spot" model with 5-degree polynomial, we just over-parametrise and get a good model for free!

This example serves as a concrete illustration of how classical statistical intuitions about model complexity may not apply in contemporary machine learning settings, particularly when dealing with over-parameterized models that have become increasingly common in practice.

\subsection{Neural Network Regression}\label{sec:nn-regression}

\cite{young1977bayesian} uses a prior that decreases with the degree of $x$. He uses orthogonal polynomials as basis functions and defines the the likelihood as
\[
P(D\mid \boldsymbol{\theta}, \sigma^2) = (2\pi\sigma^2)^{-N/2}\exp\left(-\dfrac{1}{2\sigma^2}(y-A\boldsymbol{\theta})^T(y-A\boldsymbol{\theta})\right).
\]
Here $A_{ij} = \phi_j(x_i)$. The prior is $\boldsymbol{\theta} \sim N(0,C)$, where, for example, 
$$
C =\mathrm{diag}(\delta^2,\tau^2/\lambda_1^2,\ldots,\tau^2/\lambda_q^2).
$$ 
The posterior is $\boldsymbol{\theta}\mid D, \sigma^2, C \sim N(\hat{\boldsymbol{\theta}}_{\text{post}},\Sigma_{\text{post}})$, where 
\[
\Sigma_{\text{post}} = \sigma^2(A^TA/\sigma^2 + C^{-1})^{-1}, \quad \hat{\boldsymbol{\theta}}_{\text{post}} = (A^TA/\sigma^2 + C^{-1})^{-1}A^Ty/\sigma^2.
\]

\cite{deaton1980empirical} considers the polynomial fit 
$$
y_i = P(x_i)+\epsilon_i, \quad \epsilon_i\sim N(0,\sigma^2) \; \; {\rm where} \; \; P(x) = \sum_{j=0}^m\boldsymbol{\theta}_j\psi_j(x)
$$ 
\cite{mackay1992bayesian} discusses why Legendre basis functions are better than Hermite polynomials. 
Polynomials $\psi_j(x)$ are orthogonal with respect to the data points $x_k$, such that $\sum_{k=1}^{N} \psi_i(x_k)\psi_j(x_k) = \delta_{ij}$ (assuming appropriate normalization, so $Q^TQ=I$ where $Q_{ki} = \psi_i(x_k)$).

The prior specification is given by a generalized ridge specification: $\boldsymbol{\theta}_j\sim N(0,\sigma_j^2)$.
The likelihood $Y\mid \boldsymbol{\theta} \sim N(Q\boldsymbol{\theta},\sigma^2I)$ is 
$$
P(y\mid \boldsymbol{\theta}, \sigma^2) = (2\pi\sigma^2)^{-N/2}\exp\left(-\dfrac{1}{2\sigma^2}(Y-Q\boldsymbol{\theta})^T(Y-Q\boldsymbol{\theta})\right).
$$
The OLS estimates are $\hat{\boldsymbol{\theta}}_{\text{OLS}} = Q^TY$, and $\mathrm{var}(\hat{\boldsymbol{\theta}}_{\text{OLS}}) = \sigma^2I$ as $ Q^TQ=I $. 

Using the Bayes rule for the normal-normal model (prior $\boldsymbol{\theta}_i \sim N(0, \sigma_i^2)$, likelihood from $\hat{\boldsymbol{\theta}}_i$), the posterior for $\boldsymbol{\theta}_i$ is:
\[
    \boldsymbol{\theta}_i \mid \hat{\boldsymbol{\theta}}_i,\sigma^2,\sigma_i^2 \sim N\left(\frac{\sigma_i^2}{\sigma_i^2+\sigma^2}\hat{\boldsymbol{\theta}}_i, \frac{\sigma_i^2\sigma^2}{\sigma_i^2+\sigma^2}\right).
\]
The posterior mean is $(1-z_i)\hat{\boldsymbol{\theta}}_i$ with variance $(1-z_i)\sigma^2 = z_i\sigma_i^2$ where $ z_i = \sigma^2 / (\sigma_i^2 + \sigma^2)$

Let the RSS be $s = (Y-Q\hat{\boldsymbol{\theta}}_{\text{OLS}})^T(Y-Q\hat{\boldsymbol{\theta}}_{\text{OLS}})$, then $s/\sigma^2 \sim \chi^2_{N-(p+1)}$.

The marginal distribution for $\hat{\boldsymbol{\theta}}_i$ (integrating out $\boldsymbol{\theta}_i$) is 
\[
\hat{\boldsymbol{\theta}}_i \mid \sigma^2, \sigma_i^2 \sim N(0, \sigma^2 + \sigma_i^2) = N(0, \sigma^2/z_i).
\]
Given $d = N-(p+1)$, the joint marginal likelihood of hyper-parameters $(\{z_i\}_{i=0}^m, \sigma^2)$ given data $(\{\hat{\boldsymbol{\theta}}_i\}_{i=0}^p, s)$ is proportional to (from \cite{deaton1980empirical}, Eq. 2.6, adapted):
\begin{equation}\label{eq:joint}
    L(\{z_i\}, \sigma^2 \mid \hat{\boldsymbol{\theta}},s) \propto (\sigma^2)^{-N/2} s^{(d-2)/2} \exp\left(-\dfrac{s}{2\sigma^2}\right)\prod_{i=0}^{p} z_i^{1/2} \exp\left(-\dfrac{\hat{\boldsymbol{\theta}}_i^2z_i}{2\sigma^2}\right).	
\end{equation}
Now, the question is how to estimate $z_i$ (and $\sigma^2$) from data. It is not assumed that $z_i$ are all equal. Deaton assumes an ordering:
$\sigma_0^2 \ge \sigma_1^2 \ge \ldots \ge \sigma_p^2$, which implies $0 < z_0 \le z_1 \le \ldots \le z_p \le 1$.
Deaton uses a clever  reparameterization
$$
V_i = z_i/\sigma^2 = 1/(\sigma_i^2+\sigma^2) \; \; {\rm for} \; \; i=0,\dots,p \; \; {\rm and} \; \; V_{p+1}=1/\sigma^2
$$
The constraint then becomes $V_0 \le V_1 \le \dots \le V_p \le V_{p+1}$.

The log-likelihood in terms of $V_i$ (adapted from Deaton, Eq. 2.8):
\[
\log L = \text{const} + \frac{d-2}{2}\log s + \frac{N-(p+1)}{2}\log V_{p+1} - \frac{1}{2}V_{p+1}s + \sum_{i=0}^{p}\left(\frac{1}{2}\log \frac{V_i}{V_{p+1}} - \frac{1}{2}\hat{\boldsymbol{\theta}}_i^2V_i\right).
\]
Deaton then adds priors for $V_i$. For example, Gamma priors $p(V_i) \propto V_i^{\gamma_i-1}e^{-V_i/\beta_i}$.
The unconstrained MAP estimates for $V_i$ (from differentiating the log-posterior) are of the form $\hat{V}_i^{\text{unc}} = (\text{shape}_i-1)/(\text{scale}_i + \hat{\boldsymbol{\theta}}_i^2/2)$.
These unconstrained estimates are then adjusted using isotonic regression (e.g., PAVA) to satisfy the order constraint $V_0 \le \dots \le V_p \le V_{p+1}$.
The PAVA formula is 
$$V_j^* = \max_{k \le j} \min_{l \ge j} \mathrm{Av}(k,l) \; \; {\rm where} \; \; \mathrm{Av}(k,l) = \frac{\sum_{r=k}^l w_r \hat{V}_r^{\text{unc}}}{\sum_{r=k}^l w_r}
$$
with appropriate weights $w_r$.
The condition for positive estimates often involves shape parameters $\gamma_i > 1/2$.

\paragraph{Estimation of $\sigma^2$} \cite{deaton1980empirical} provides an empirical Bayes approach where $V_{p+1} = 1/\sigma^2$ is estimated as part of the isotonic procedure.

To illustrate Deaton's approach, consider the case when the true model is as follows. We use normalized Legendre polynomials as the basis:
\begin{align*}
y(x, \boldsymbol{\theta}) = f(x, \boldsymbol{\theta}) + \epsilon, \quad \epsilon_i \sim N(0,0.3^2).
\end{align*}
In matrix form, for a vector of inputs $\mathbf{x} = [x_1, x_2, \ldots, x_N]^T$, with $N=20$:
\[
    \mathbf{y} = \mathbf{Q}\boldsymbol{\theta} + \boldsymbol{\epsilon}.
\]
\begin{figure}[H]
	\centering
	\includegraphics[width=0.6\textwidth]{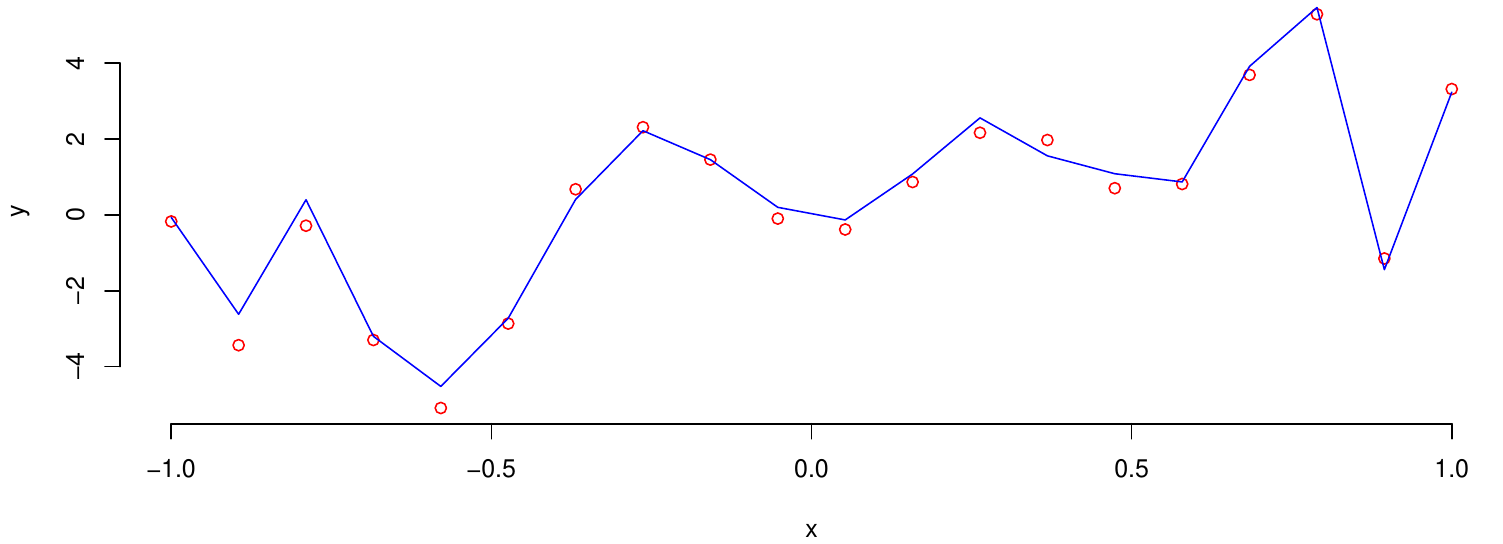}
    \caption{Data generated from the model $y(x, \boldsymbol{\theta}) = f(x, \boldsymbol{\theta}) + \epsilon$, where $\epsilon_i \sim N(0,0.3^2)$. The red dots represent the true function $f(x, \boldsymbol{\theta})$, and the blue line shows the noisy observations.}
	\label{fig:deaton-data}
\end{figure}

\begin{figure}[H]
    \includegraphics[width=0.6\textwidth]{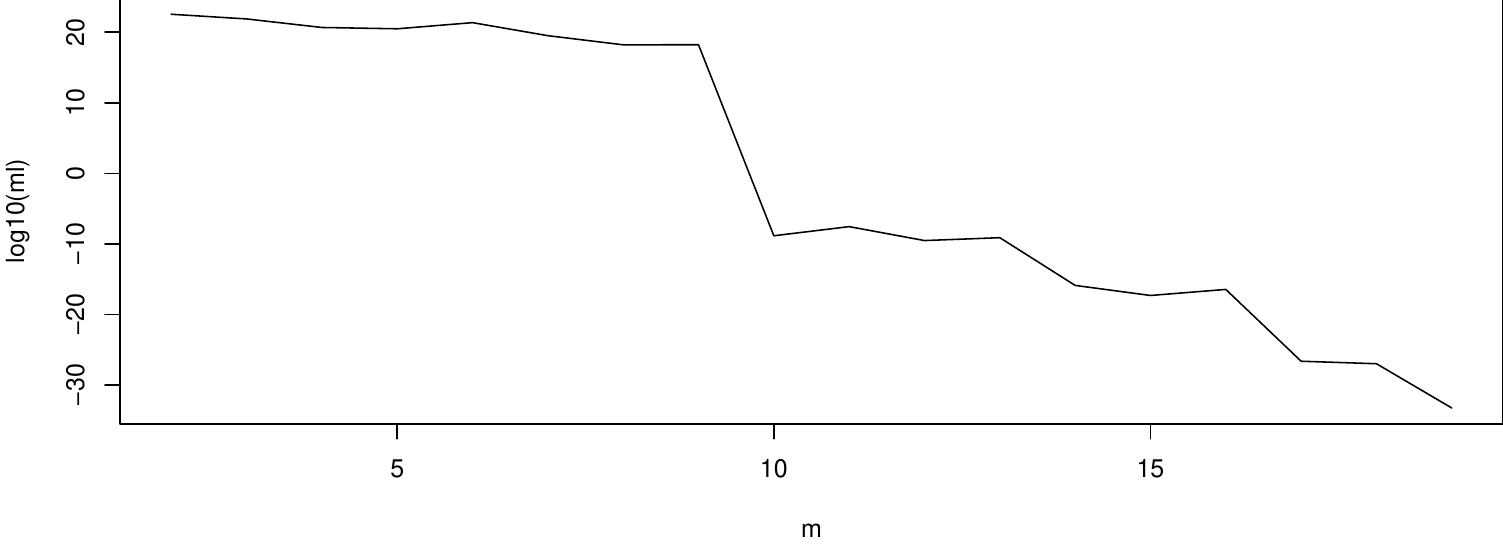}
    \caption{Marginal likelihood for a model with true polynomial degree $p_{\text{true}}=10$ (example) and $N=20$ observations. The x-axis represents the assumed model complexity $p$ (degree of polynomial fit), and the y-axis shows the log marginal likelihood. The peak indicates the optimal model complexity as selected by the marginal likelihood.}
    \label{fig:deaton-marginal}
\end{figure}

Figure~\ref{fig:deaton-marginal} shows the marginal likelihood on a logarithmic scale for different model complexities $m$. 

Having established the Bayesian framework for polynomial regression and neural networks, we now explore how these ideas can be extended to more flexible basis functions that can adapt to the data structure.

\paragraph{Neural Network Basis}
The Bayesian framework naturally extends to the neural network case. Let $\phi_j(x; \mathbf{w})$ denote the $j$-th neural network basis function with parameters $\mathbf{w}$. The regression model becomes:
\[
y_i = \sum_{j=1}^M \boldsymbol{\theta}_j \phi_j(x_i; \mathbf{w}) + \epsilon_i, \quad \epsilon_i \sim N(0, \sigma^2)
\]

The hierarchical Bayesian specification includes priors on both the basis function parameters $\mathbf{w}$ and the linear combination coefficients $\boldsymbol{\theta}$:
\begin{align}
\boldsymbol{\theta}_j &\sim N(0, \sigma_j^2), \quad j = 1, \ldots, M\\
\mathbf{w} &\sim p(\mathbf{w})\\
\sigma_j^2 &\sim p(\sigma_j^2)
\end{align}

This allows for adaptive learning of basis functions while maintaining the Bayesian model selection framework. The marginal likelihood can still be computed (approximately) to select the optimal number of basis functions $M$, potentially exhibiting the double descent phenomenon as $M$ varies.

\paragraph{Computational Considerations}
Extending the Bayesian framework to neural network basis functions introduces several computational challenges. Maintaining orthogonality among the learned basis functions during training requires additional mechanisms, such as projection methods or the inclusion of penalty terms in the loss function. Furthermore, the computation of the marginal likelihood becomes substantially more complex, owing to the nonlinear dependence on the neural network parameters $\mathbf{w}$. As a result, standard analytical approaches are often infeasible, and approximate inference techniques—such as variational inference or Markov chain Monte Carlo sampling—may be necessary to perform posterior computations in this setting.

Despite these challenges, neural network basis functions offer several notable advantages over fixed polynomial bases. Most importantly, they provide adaptivity, allowing the basis functions to conform to the specific structure present in the data. Neural networks also possess greater representational power, enabling them to capture more complex patterns than traditional polynomial expansions. In addition, neural networks can automatically learn relevant features from the data, eliminating the need to pre-specify polynomial orders or other basis parameters. Finally, this approach is potentially more scalable, as neural networks are well-suited to handling high-dimensional input spaces where polynomial expansions would become unwieldy.

This neural network extension provides a promising direction for combining the interpretability and theoretical guarantees of Bayesian polynomial regression with the representational power of deep learning, while maintaining the double descent framework we have developed.

The connection between neural networks and ridge regression becomes clearer when we consider the canonical basis representation. In the following section, we develop the theoretical foundations of generalized ridge regression and show how it relates to the Bayesian double descent phenomenon we have been studying.

\section{Discussion}\label{sec:discussion}

This paper has provided a comprehensive Bayesian perspective on the double descent phenomenon. We have shown that double descent is naturally understood from a Bayesian viewpoint, where there is a natural bias-variance trade-off built into the Bayesian paradigm. This also leads to an automatic Occam's razor---the automatic penalization of model complexity via the marginal likelihood.

The Bayesian paradigm provides a coherent framework to simultaneously infer parameters and model complexity, and thus to understand double descent, namely the re-descending property of Bayes risk as a function of complexity. The optimal predictive rule is adaptive and is a weighted average of individual predictors given complexity.

In this work, we have made several key contributions to the understanding of double descent from a Bayesian perspective. First, we have shown that the conditional prior specification $p(\boldsymbol{\theta}_M | M)$ serves as an implicit regularizer, fundamentally driving the double descent phenomenon. This insight provides a natural Bayesian interpretation of double descent, filling a gap that previously existed in the literature.

Second, we have clarified the relationship between Bayesian Occam's razor and the double descent effect, demonstrating that the perceived conflict between the two is, in fact, illusory. While the marginal likelihood in Bayesian model selection tends to favor simpler models, the conditional prior can still yield favorable risk properties even in highly over-parameterized regimes.

Third, we have established a comprehensive theoretical foundation for these ideas, drawing on Dawid's model comparison theory, the Dickey-Savage results, and connections to generalized ridge regression and shrinkage methods. These theoretical developments provide a rigorous basis for understanding the interplay between prior specification, model complexity, and predictive risk.

Finally, we have illustrated how the Bayesian framework naturally extends to modern machine learning methods, such as neural networks, offering new insights into their generalization properties. This connection underscores the broad applicability of the Bayesian approach to contemporary challenges in statistical learning.

There are several important implications for cross-validation and the choice of regularization. This highlights the trade-off between training and test error in neural networks and other over-parameterized models. The Bayesian approach suggests that instead of relying solely on cross-validation for model selection, practitioners should also consider the implicit regularization provided by appropriate prior specifications.

There are many areas for future research. In higher dimensions, see the discussion in~\cite{polson2017deep} where one can find an "island of good estimators." Counterintuitively, this can be in the opposite direction of the classical bias-variance trade-off. Variance can be minimized by appealing to exchangeability results due to Kingman~\cite{kingman1978uses} for exchangeable variables. See \cite{amit1997shape} for the discrete classification case and minimization of entropy of misclassification.

Fully eliciting priors and understanding their relationship with risk characteristics are still active areas of research. The connection between global-local shrinkage priors and double descent deserves further investigation, particularly in the context of deep learning where the number of parameters can be orders of magnitude larger than the sample size.

Another promising direction is the development of adaptive prior specifications that can automatically adjust to the data structure, potentially leading to more robust double descent behavior across different problem domains.

\bibliography{DoubleDescent} 

\appendix
\section{Asymptotic Posterior Model Probabilities}

This section follows \cite{berk1966limiting} and \cite{philipdawid2011posterior}. The behavior of the posterior when the true model is not in the model set under consideration is also of interest.

\paragraph{Asymptotic carrier} The posterior $p(\boldsymbol{\theta} | y, M)$ has an optimality property when $ f_{true} \notin M $. This is quantified by the \emph{asymptotic carrier} of the posterior, $\mathcal{C}$, which is defined by:
\[
\mathcal{C} = \argmin_{\boldsymbol{\theta} \in \Theta_M} \int f_{\text{true}}(y) \log \frac{f_{\text{true}}(y)}{f(y|\boldsymbol{\theta},M)} dy = \argmin_{\boldsymbol{\theta} \in \Theta_M} \mathrm{KL}(f_{\text{true}} || f_{\cdot|\boldsymbol{\theta},M}).
\]
That is, the posterior over parameters in the model class $\mathcal{M}$ (for a given $M$) converges to the density $f(\cdot|\boldsymbol{\theta}^*, M)$ where $\boldsymbol{\theta}^*$ minimizes the Kullback-Leibler (KL) divergence between the data generating process $f_{\text{true}}$ and the model class. More precisely, for any neighborhood $\mathcal{A}$ of $\mathcal{C}$, we have:
\[
\lim_{n \rightarrow \infty} P[\boldsymbol{\theta} \in \mathcal{A} | y_1, \ldots, y_n, M] = 1 \quad \text{a.s. } F_{\text{true}}.
\]
Since the initial paper of~\cite{schwartz1965bayes}, there has been much work on the limiting behavior of the posterior when the true model $f_{\text{true}}$ lies in the class $f_{\boldsymbol{\theta}}$ indexed by models. \cite{berk1966limiting} provides an interesting result on the posterior when the model is ``incorrect'' (a.k.a. we only calculate posterior probabilities over models in our given class). Bayes is coherent, and in this sense, we can only calculate relative posterior probabilities. We can simply increase the space of models and re-calculate. Lindley shows the rationality of this approach and terms this process ``the art of extending the conversation.'' See also the open and closed model framework of~\cite{bernardo2000bayesian}. Convergence in model space to the true model can happen very quickly---at an exponential rate.

Let $M \in \mathcal{M}$ and associated with the model $\boldsymbol{\theta}_M \in \Theta_M$. When model $M$ attains then parameter $\boldsymbol{\theta}_M \sim \Pi_M(\boldsymbol{\theta}_M)$ with respect to Lebesgue measure $d\boldsymbol{\theta}_M$ over $\Theta_M$.

Suppose that data $Y \sim Q$. Then, if $K(Q, M_2) > K(Q, M_1)$ where $K$ is the min KL divergence between $Q$ and models in $M$, the log-Bayes factor is given by:
\[
\log \frac{p(Y|M_1)}{p(Y|M_2)} \sim n(KL(Q, M_2) - KL(Q, M_1)) + O_P(n^{1/2})
\]

If $Q \in M_1 \cap M_2$, then $KL(Q, M_2) = KL(Q, M_1)$ and we have the asymptotic:
\[
\log \frac{p(Y|M_1)}{p(Y|M_2)} = \frac{1}{2}(d_{M_2} - d_{M_1}) \log \left(\frac{n}{2\pi e}\right) + \log \frac{\rho(\boldsymbol{\theta}_1^\star | M_2)}{\rho(\boldsymbol{\theta}_1^\star | M_1)} + O_P(1)
\]
where $\rho(\boldsymbol{\theta} | M) = \pi_M(\boldsymbol{\theta}) |\det I_M(\boldsymbol{\theta})|^{-1/2}$ is an Occam factor. See also Bartlett (1941) paradox and the asymptotic behavior of Bayes factors.

Having established the Bayesian framework for model complexity and the role of priors in model selection, we now turn to the main contribution of this paper: understanding double descent from a Bayesian perspective. The key insight is that the conditional prior specification $p(\boldsymbol{\theta}_M | M)$ acts as an implicit regularizer that can lead to the re-descending risk behavior characteristic of double descent. This connects the theoretical foundations we have developed---including Dawid's model comparison theory, the Dickey-Savage density ratio, and the BIC approximation---to the empirical phenomenon of double descent.
\begin{proof}
We can write
\begin{equation}
\log \frac{p_n(X_n)}{q_n(X_n)} = \log \frac{p_n(X_n)}{p_n(X_n|\hat{\omega}_n)} + \log \frac{p_n(X_n|\hat{\omega}_n)}{p_n(X_n|\omega^*)} + \log \frac{p_n(X_n|\omega^*)}{q_n(X_n)}
= A + B + C \text{ say,}
\label{eq:A1}
\end{equation}
where $\hat{\omega}_n$ denotes the maximum likelihood estimator of $\omega$ in model $M$ based on data $X_n$. Henceforth we drop $n$ from the notation.

Note that the prior distribution only enters into the term $A$. The asymptotic form of the term $A$, based on Laplace expansion, is well-known \citep{tierney1986accurate}. We have:
\begin{equation}
A = \frac{1}{2}d \log 2\pi + \log \frac{p(\hat{\omega})}{\{\det -l''(\hat{\omega})\}^{1/2}} + O(n^{-1}).
\label{eq:A2}
\end{equation}
This expansion is essentially algebraic, rather than probabilistic, and will typically hold for essentially all data sequences, whether or not they appear to be generated by some distribution in the model $M$; in particular, we can regard the final term as $O_p(n^{-1})$ under $Q$.

Under $Q$, $\hat{\omega} \to \omega^*$ almost surely \citep{wald1949note}. Taylor expansion in the neighbourhood of $\hat{\omega}$ yields:
\begin{equation}
l'(\omega) \approx l''(\hat{\omega})(\omega - \hat{\omega}),
\label{eq:A3}
\end{equation}
so that
\begin{equation}
\hat{\omega} - \omega \approx -l''(\hat{\omega})^{-1} l'(\omega).
\label{eq:A4}
\end{equation}

Under $Q$, each of $l'(\omega)$, $l''(\omega)$ is a sum of $n$ independent and identically distributed components; moreover, it may be checked that $E_Q\{l'(\omega^*)\} = 0$. It follows that
\begin{equation}
n^{1/2}(\hat{\omega} - \omega^*) \xrightarrow{L} N\{0, (J_*^2)^{-1})^T J_*^1 J_*^{2^{-1}}\},
\label{eq:A5}
\end{equation}
where $J_*^1 = J_1(\omega^*)$, $J_*^2 = J_2(\omega^*)$, with $J_1(\omega) = \text{var}_Q\{l'_1(\omega)\}$, $J_2(\omega) = E_Q\{-l''_1(\omega)\}$.

In particular, under $Q$, $\hat{\omega} = \omega^* + O_p(n^{-1/2})$, $-l''(\hat{\omega}) = nJ_*^2 + O_p(n^{1/2})$. It then follows from \eqref{eq:A2} that
\begin{equation}
A = -\frac{1}{2}d \log \frac{n}{2\pi} + \log \frac{p(\omega^*)}{\{\det J_*^2\}^{1/2}} + O_p(n^{-1/2}).
\label{eq:A6}
\end{equation}
Note in particular that the asymptotic effect of the within-model prior density $p(\omega)$ on $\log \frac{p_n(X_n)}{q_n(X_n)}$ is entirely captured in its order 1 contribution to \eqref{eq:A6}.

From this point on we consider the two cases (i) and (ii) individually.

\begin{enumerate}
\item[(i)] If $Q \not\in M$, then $K(Q, M) > 0$. In this case, under $Q$, $\log p(X|\omega^*)/q(X)$ is a sum of $n$ independent and identically distributed components with mean $-K(Q, M)$, and thus term $C$ in \eqref{eq:A1} has the form
\begin{equation}
C = -nK(Q, M) + \left[n \text{var}_Q\left\{\log \frac{p_1(X|\omega^*)}{q_1(X)}\right\}\right]^{1/2} Z,
\label{eq:A7}
\end{equation}
where $Z \xrightarrow{L} N(0, 1)$. Equation \eqref{eq:A6} shows that $A = O_p(\log n)$. Finally, consider $B = l(\hat{\omega}) - l(\omega^*)$. By Taylor expansion,
\begin{align}
B &\approx \frac{1}{2}\{n^{1/2}(\hat{\omega} - \omega^*)\}^T \{-n^{-1}l''(\hat{\omega})\}\{n^{1/2}(\hat{\omega} - \omega^*)\} \nonumber \\
&\approx \frac{1}{2}\{n^{1/2}(\hat{\omega} - \omega^*)\}^T \{J_*^2\}\{n^{1/2}(\hat{\omega} - \omega^*)\},
\label{eq:A8}
\end{align}
which, by \eqref{eq:A5}, is $O_p(1)$. Thus both terms $A$ and $B$ are of smaller order than $C$, and (4) follows.

\item[(ii)] Now suppose $Q \in M$, so that $K(Q, M) = 0$ and $Q = P_{\omega^*}$. In this case, term $C$ in \eqref{eq:A1} vanishes. We also now have $J_*^1 = J_*^2 = I(\omega^*)$. Thus
\begin{equation}
A = -\frac{1}{2}d \log \frac{n}{2\pi} + \log \vartheta(\omega^*) + O_p(n^{-1/2}).
\label{eq:A9}
\end{equation}
Also by \eqref{eq:A8} and \eqref{eq:A5}, or directly from Wilks' Theorem,
\begin{equation}
B \xrightarrow{L} \frac{1}{2}\chi_d^2,
\label{eq:A10}
\end{equation}
and in fact the distribution of $B$ will typically differ from that of $\frac{1}{2}\chi_d^2$ by $O(n^{-1/2}$. Then (5) follows. (For a more rigorous treatment, see \citet{Clarke1990}.)
\end{enumerate}
\end{proof}

\end{document}